\documentclass[10 pt, conference]{ieeeconf}  

\IEEEoverridecommandlockouts                              

\usepackage{amssymb,url,amsmath,bm}
\usepackage[ruled,vlined]{algorithm2e}
\usepackage{graphicx}
\graphicspath{ {Images/} }
\usepackage{epstopdf}
\usepackage{textcomp}
\usepackage{color}
\usepackage{ifpdf}
\usepackage{url}
\usepackage{enumerate}
\usepackage{array}
\usepackage{flushend}
\usepackage{multirow,booktabs}
\usepackage[pdfstartview=]{hyperref}
\usepackage[export]{adjustbox} 
\usepackage{caption}
\usepackage{subcaption}
\usepackage{mdwlist}
\usepackage[utf8]{inputenc}
\usepackage{todonotes}

\usepackage{tcolorbox}
\newtcolorbox[auto counter]{algorithmbox}[2][]{colback=red!5!white,colframe=red!75!black,fonttitle=\bfseries, title=\alg\thetcbcounter: #2,#1}

\newcommand{\alg}[1]{Algorithm~\ref{#1}}

\newcommand{\fig}[1]{Fig.~\ref{#1}}

\usepackage{cite}

\title{Direct Mutation and Crossover in Genetic Algorithms Applied to Reinforcement Learning Tasks}

\author{Tarek Faycal$^{1}$ and Claudio Zito$^{2\ast}$
\thanks{$^{1}$IRLab, School of Computer Science, University of Birmingham, United Kingdom}%
\thanks{$^{2}$Technology Innovation Institute, Abu Dhabi, UAE}
\thanks{$^{\ast}$Correspondent author
{\tt\small Claudio.Zito@tii.ae}}
}
\date{}

\begin{document}

\maketitle

\begin{abstract}

Neuroevolution has recently been shown to be quite competitive in reinforcement learning (RL) settings, and is able to alleviate some of the drawbacks of gradient-based approaches. This paper will focus on applying neuroevolution using a simple genetic algorithm (GA) to find the weights of a neural network that produce optimally behaving agents. In addition, we present two novel modifications that improve the data efficiency and speed of convergence when compared to the initial implementation. The modifications are evaluated on the FrozenLake environment provided by OpenAI gym and prove to be significantly better than the baseline approach.

\end{abstract}

\section{Introduction}
\label{S:1}

Neuroevolution is the subset of artificial intelligence techniques that makes use of evolutionary algorithms to determine the topology and parameters of neural networks, inspired by natural evolution. Evolutionary methods solve optimization problems in a population-based, generate-and-test fashion.This approach is quite different from the traditional Reinforcement Learning (RL) setting, where a single agent interacts with the environment and its performance is evaluated over episodes rather than generations of solutions, as in our previous work \cite{faycal2022dynat}. This type of search can be insensitive to delayed rewards because the fitness does not necessarily depend on the environment's reward signal. This evolutionary approach has been shown to be successful on its own, like \cite{such2017deep},  \cite{salimans2017evolution}, and \cite{igel2003neuroevolution}. In combination with traditional RL, it has proven to be useful in finding optimal hyperparameters \cite{DBLP:journals/corr/abs-1711-09846}, and in population-based training of RL agents to master complex games faster like Capture the Flag \cite{Jaderberg859} or StarCraft \cite{10.1038/s41586-019-1724-z}.

In this paper, we present a baseline Genetic Algorithm, that uses mutation and crossover to evolve a densely connected feedforward network. Following that, we present modifications to both the mutation and crossover operators that are tested on the classic RL environment, FrozenLake. Both modifications present a significant improvement over using the baseline.

\section{Theoretical Background}
\label{S:2}

\begin{figure*}[t]
       \centering
       \captionsetup{justification=centering}
       \includegraphics[scale=0.45]{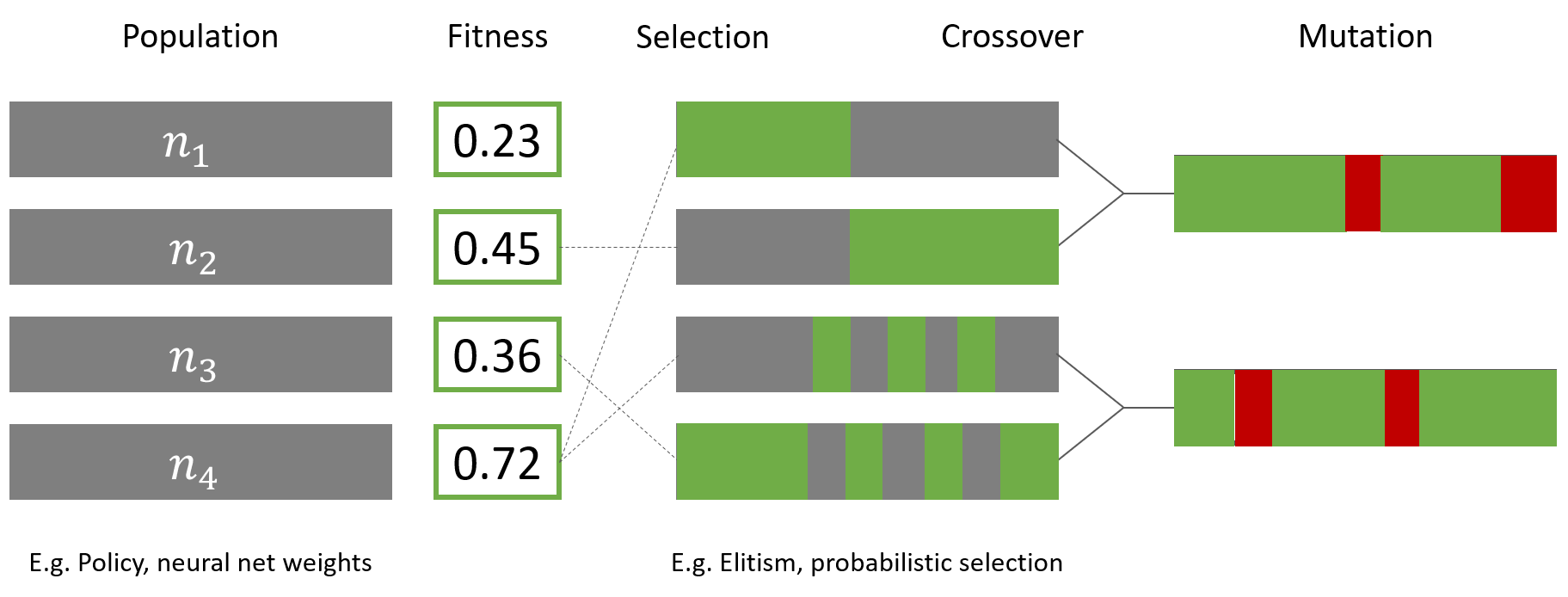}
       \caption[Genetic Algorithm Illustration] {Genetic Algorithm Illustration: A population is generated and evaluated, the elite performers are then selected for crossover and mutation to generate a new, more fit population}
       \label{fig:ga}
\end{figure*}

Genetic Algorithms (GA) \cite{holland1975adaptation} are a class of evolutionary algorithms that make use of biology-inspired mechanisms such as crossover, mutation, and selection. They can be considered as a robust general-purpose optimization technique; commonly used, for example, to solve the inverse kinematics problem for redundant robots, e.g., \cite{bib:zito_2016, bib:zito_w2012, bib:zito_w2013, bib:rosales_2018, bib:zito_2019}.The implementation used starts by generating a population, weights of a neural network in this case, and evaluating how well each individual performs in the environment. The evaluation will result in a fitness score that is used to rank individuals according to fitness. To generate the next population, the \textbf{elite} performers are carried over unchanged to guarantee either an increase or level performance across generations, this is the \textbf{selection} portion of the algorithm. The elite performers are then crossed over by randomly choosing two ``parent'' individuals and selecting traits from both with equal probability to pass on to the ``child''. The resulting population of neural network weights is mutated by applying Gaussian noise drawn from the same distribution to all individuals in the population. An illustration of this process is shown in \fig{fig:ga}.

Crossover is usually done in a uniformly random manner, which does not guarantee good traits are passed on to children, making it less likely that a suitable solution will be generated. We attempt to remedy this with Directed Crossover. 
In the baseline, fitness is simply the reward gathered in the environment. In Novely Search \cite{lehman2011abandoning}, fitness does not consider external reward at all, and fitness is how different generated solutions are from previous ones. We present a method that makes use of both these ideas with Multi-step Mutation. To support this claim, ten samples of each algorithm are collected and the median scores are reported, differences in performance are statistically significant. However, due to computational constraints this approach was not tested in a parallelized manner on a large scale environment, nor was it tested with different network architectures. Further experimentation would be needed to validate performance on harder problems.  

Our main claim is that our modifications increase the speed of convergence, which is seen as a reduction in the number of generations needed to find a successful candidate solution.

\section{Related Work}
\label{S:3}

The application of neuroevolution methods to RL is not a new phenomenon (\cite{whiteson2006evolutionary}, \cite{zhen2013neuroevolution}, \cite{igel2003neuroevolution}), but they were only recently shown to be competitive with the Deep RL approach in \cite{salimans2017evolution}, and \cite{such2017deep}. In the aforementioned papers, evolutionary methods were used, Natural Evolution Strategies in \cite{salimans2017evolution}, and a Genetic Algorithm in \cite{such2017deep}, to determine the parameters of neural networks to play Atari games, and for agent locomotion in a simulated physical environment. In both papers, these approaches yield results that are comparable to Deep RL, and even better in a few cases. One advantage is that evolutionary methods are highly parallelizable, since evaluation does not need to be carried out in a sequential fashion, meaning they do provide a significant speedup when it comes to learning. Unfortunately, this parallelization requires access to a lot of computational power to realize this advantage. For example, in \cite{salimans2017evolution} the task of 3D humanoid locomotion was evaluated at different numbers of CPU cores, and reaching a predetermined score takes ~11 hours using 18 cores, and 10 minutes using 1440 cores. The speedup is significant but quite costly.

The neuroevolution approach adopted in this project uses a simple Genetic Algorithm that generates a population of weights, tests the population, and carries out crossover and mutation on a portion of the population that is successful to produce the next generation. A small number of top performers are moved to the next generation unchanged to preserve the best solutions, this approach is known as elitism. The reason behind this choice is that there is no gradient estimation or calculation happening at any point in the process (as opposed to Deep RL and Evolution Strategies), freeing it from any problems that accompany that approach.

\section{Problem Formulation}
\label{S:4}

Our aim in this paper is to evolve a feedforward network using a Genetic Algorithm and apply it to a reinforcement learning task. It's been shown to produce architectures comparable with the best hand-designed methods. For our implementation of the GA, crossover was done uniformly, by selecting two elite performers, and randomly choosing weights from each until a the child network is complete. Mutation was applied as noise sampled from a normal distribution in the shape of the network's weight matrix. 

\begin{figure*}[t]
       \centering
       \captionsetup{justification=centering}
       \includegraphics[scale=0.6]{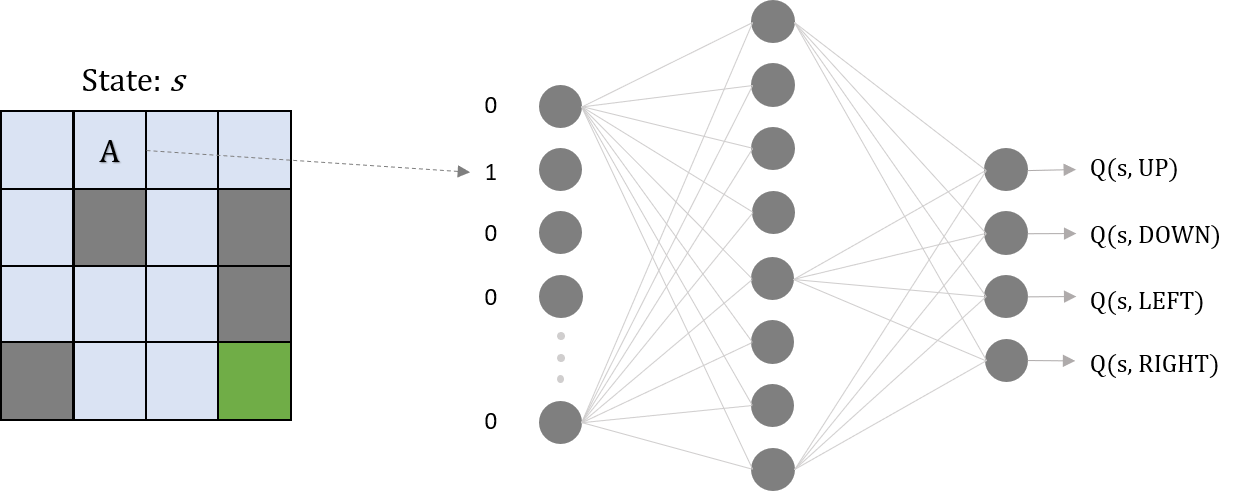}
       \caption[Neuroevolution network setup] {Neuroevolution network setup: The state is fed in as a one-hot vector and the output is the Q-value associated with each action. The action the agent takes is the one with the largest value.\textbf{A} is the agent's position in this example} 
       \label{fig:frozenlake}
\end{figure*}

In \cite{such2017deep}, the fitness function did not reward the networks' direct performance on the game in \fig{fig:frozenlake}. They use a method called Novelty Search \cite{lehman2011abandoning}, that rewards solutions that are simply different from previously generated ones. While this approach is quite successful and competitive with Deep RL and Evolution Strategies, it strays from the RL paradigm by ignoring the environment's reward signal. Our baseline implementation evaluates fitness by carrying out an episode in the environment and using the gathered rewards to represent fitness, as shown in \alg{alg:baseline}. 

\bigskip
\begin{algorithm}[t]
\DontPrintSemicolon
\SetAlgoLined
 $p \leftarrow generate\_population()$\;
 \While{not converged}{
  $f \leftarrow compute\_fitness(p)$\;
  $e \leftarrow$ elite performers\;
  $p_{new} \leftarrow mutate(crossover(e))$\;
  $p \leftarrow p_{new}$\;
 }
 \caption{Baseline GA}
 \label{alg:baseline}
\end{algorithm}
\bigskip

\subsection{Modification 1: Multi-step Mutation} 
In an effort to direct the variation-selection process carried out in GA, we present a novel modification called multi-step mutation (MSM). The method is carried out described in \alg{alg:msm}.

\bigskip
\begin{algorithm}[t]
\DontPrintSemicolon
\SetAlgoLined
 $n \leftarrow$ number of additional mutations\;
 $p \leftarrow generate\_population()$\;
 \While{not converged}{
  $f_{current} \leftarrow compute\_fitness(p)$\;
  $e \leftarrow$ elite performers\;
  $p_{mutated} \leftarrow e$\;
  \For{$i\gets0$ \KwTo $n-1$}{
    $p_{mutated} \leftarrow mutate(crossover(p_{mutated}))$\;
  }
  $f_{mutated} \leftarrow compute\_fitness(p_{mutated})$
  
  \eIf{$f_{mutated} \geqslant f_{current}$}{
    $p \leftarrow p_{mutated}$\;
   }{
   $p_{new} \leftarrow mutate(crossover(e))$\;
   $f_{new} \leftarrow compute\_fitness(p_{new}, p_{mutated})$\;
   $e_{new} \leftarrow$ select elite performers according to $f_{new}$\;
   $p \leftarrow mutate(crossover(p_{new}))$\;
  }
 }
 \caption{Multi-step Mutation}
 \label{alg:msm}
\end{algorithm}
\bigskip

\begin{figure*}[t]
       \centering
       \captionsetup{justification=centering}
       \includegraphics[scale=0.55]{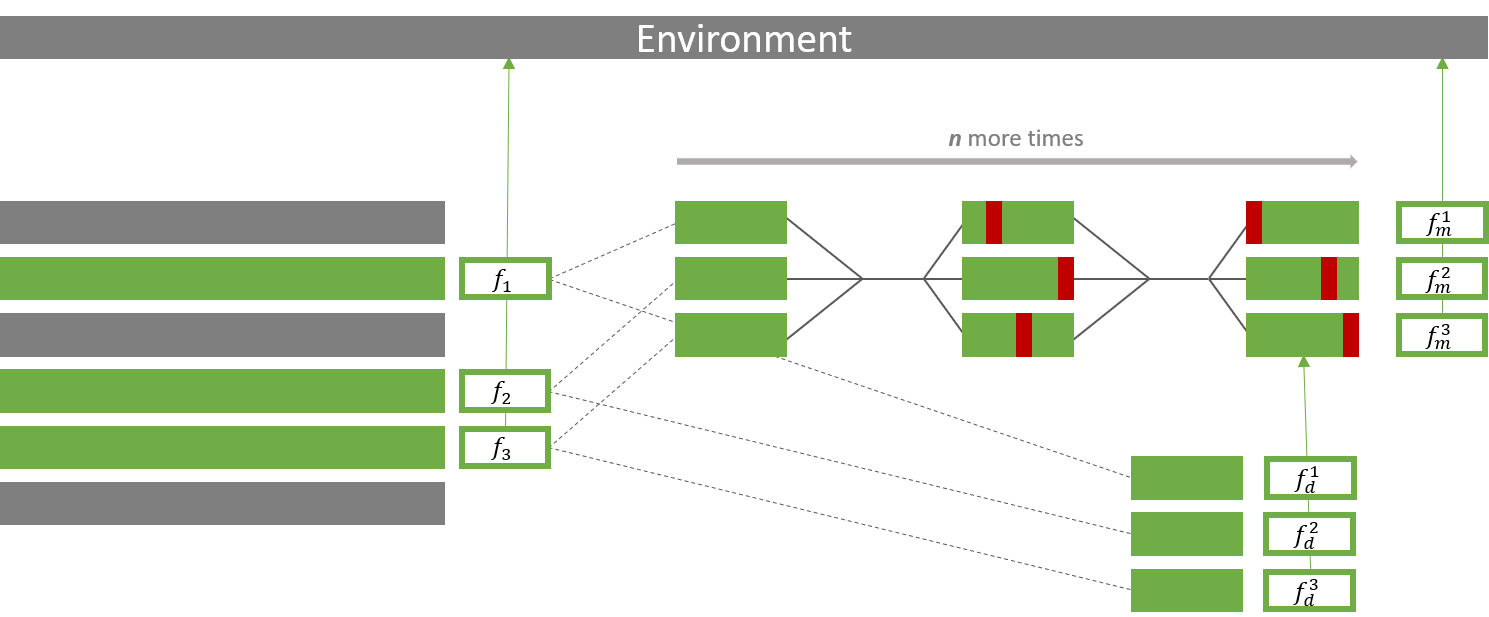}
       \caption[MSM Illustration] {MSM Illustration: The elite performers are crossed over and mutated several extra times and evaluated again. If this new population does better it is kept. Otherwise, the previous set of elites is used to generate a new population, whose fitness is evaluated as the distance from the n-step mutated population. Note that $f_{d}$ represents the distance maximizing internal fitness measure.}
       \label{fig:msm}
\end{figure*}

They key differentiation from standard GA is the added mutation and crossover steps, in addition to the condition attached to the mutated population's performance. If the mutated population does better than the current population, it is set as the new population. Otherwise, we apply crossover and mutation to the current population using dissimilarity from the mutated population as a fitness measure, as opposed to using fitness in the target domain (\fig{fig:msm}).

Intuitively we believe that the internalized fitness measure would help traverse the possible solution space faster, using only the relative fitness of generated populations in a gradient-free manner. Moreover, it allows for ``anti-goal'' generation as a result of the look-ahead process. This has the potential to make a traditionally blind variation process slightly more directed in the sense that there is now a target to avoid as well as the underlying reward function that the algorithm is trying to maximize. Going forward, it would be interesting to try and reduce the interactions needed between candidate solutions and the environment in a way that allows this optimization to be carried out in a single, structured online agent playing a game (RL paradigm) rather than black-box optimization over populations.

\subsection{Modification 2: Directed Crossover in Sparse Networks}

Crossover is usually performed in a uniform manner, with an equal probability of selecting a trait from either parent. When the traits are weights of a neural network, there is no guarantee that a random combination of weights from two parents would result in a better performing child network. Mocanu et. al. \cite{Mocanu2018SET} show that neural networks initialized with sparse connections between the layers can achieve the same accuracy as fully connected networks. They accomplish this via the Sparse Evolutionary Training procedure introduced in their paper. The connections between two layers are randomly initialized using an Erdos-Renyi random graph, and at each stage of training, a certain fraction of the weights, the ones closest to zero, are removed and new random connections are added, maintaining the same number of weights. Their results are promising in both a supervised and unsupervised setting, using Multi-Layer Perceptrons, Convolutional Neural Networks, and Restricted Boltzmann Machine architectures.

We modify this approach to fit the GA-style optimization paradigm, utilizing it specifically during parameter generation and crossover. The modified method is then tested in a reinforcement learning environment. The modifications to baseline GA are as shown in \alg{alg:direct_crossover}.
\bigskip

\begin{algorithm}[t]
\DontPrintSemicolon
\SetAlgoLined
 $p \leftarrow generate\_population()$\;
 $m \leftarrow generate\_sparsity\_masks()$\;
 $p \leftarrow p * m$\;
 \While{not converged}{
      $f \leftarrow compute\_fitness(p)$\;
      $e \leftarrow$ elite performers\;
      $p_{new} \leftarrow mutate(directed\_crossover(e))$\;
      $p \leftarrow p_{new}$\;
 }
 \caption{Directed Crossover}
 \label{alg:direct_crossover}
\end{algorithm}

\bigskip
Directed Crossover:
    \begin{enumerate}
        \item Remove a fraction $\zeta$ of the parents' weights closest to zero as they contribute the least to success
        \item Set the child's parameters to the larger of the parents' values
        \item Add random connections sampled from a Gaussian distribution to make up for the lost connections 
    \end{enumerate}

If the largest weights contribute most to the parents' success, then those traits should be the ones passed on to children. Tests show a significant improvement over the baseline approach in the environment it was tested in. Moreover, the resulting networks are sparse, which means increased memory efficiency at scale. Further modifications could include the use of more than two parents for crossover while clearing out a larger fraction of the weights close to zero.

\section{Experiments}
\label{S:5}

\subsection{Setup}

\begin{figure}[t]
       \centering
       \captionsetup{justification=centering}
       \includegraphics[scale=1.5]{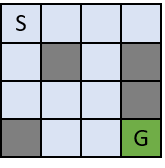}
       \caption[The FrozenLake Environment] {The FrozenLake Environment: The agent must navigate from start to finish, the grey squares are holes that terminate the episode with a reward of 0, while the blue squares are frozen and can cause an agent to slip in a random direction. The only positive reward available is 1, if the agent reaches the goal.}
       \label{fig:frozenlake_env}
\end{figure}

For this study, we use the FrozenLake environment of OpenAI Gym \cite{1606.01540} to evaluate the algorithms (\fig{fig:frozenlake_env}). FrozenLake is a tabular, stochastic, fully observable, and discrete environment that  does not approach the size of ones used in current deep reinforcement learning research, but the reduced complexity is beneficial for validating and understanding the behaviour of our algorithms.

The agent must navigate from \textbf{S} to \textbf{G}, and avoid holes, the grey squares, which terminate the episode with a reward of 0. The blue squares are frozen, meaning the agent can slip in a random direction whenever it takes an action while it's on a blue square. The available actions are Up, Down, Left, and Right. The rewards are sparse because the agent only receives a reward of 1 if it manages to reach the goal state. Due to the nature of the environment, for an algorithm 
to solve this task it must achieve an average score of 0.78, at least, over 100 episodes. 

The evaluation of neuroevolution algorithms was done as follows: Ten samples from each of the baseline, MSM, and Directed Crossover were gathered. The Mann-Whitney U score was computed on the population's mean, as well as the top performer's score at the end of 500 generations. The results are presented in the next section. 

\subsection{Results}

\begin{figure}[t]
\captionsetup{justification=centering}
\begin{subfigure}{0.5\textwidth}
\includegraphics[width=\linewidth]{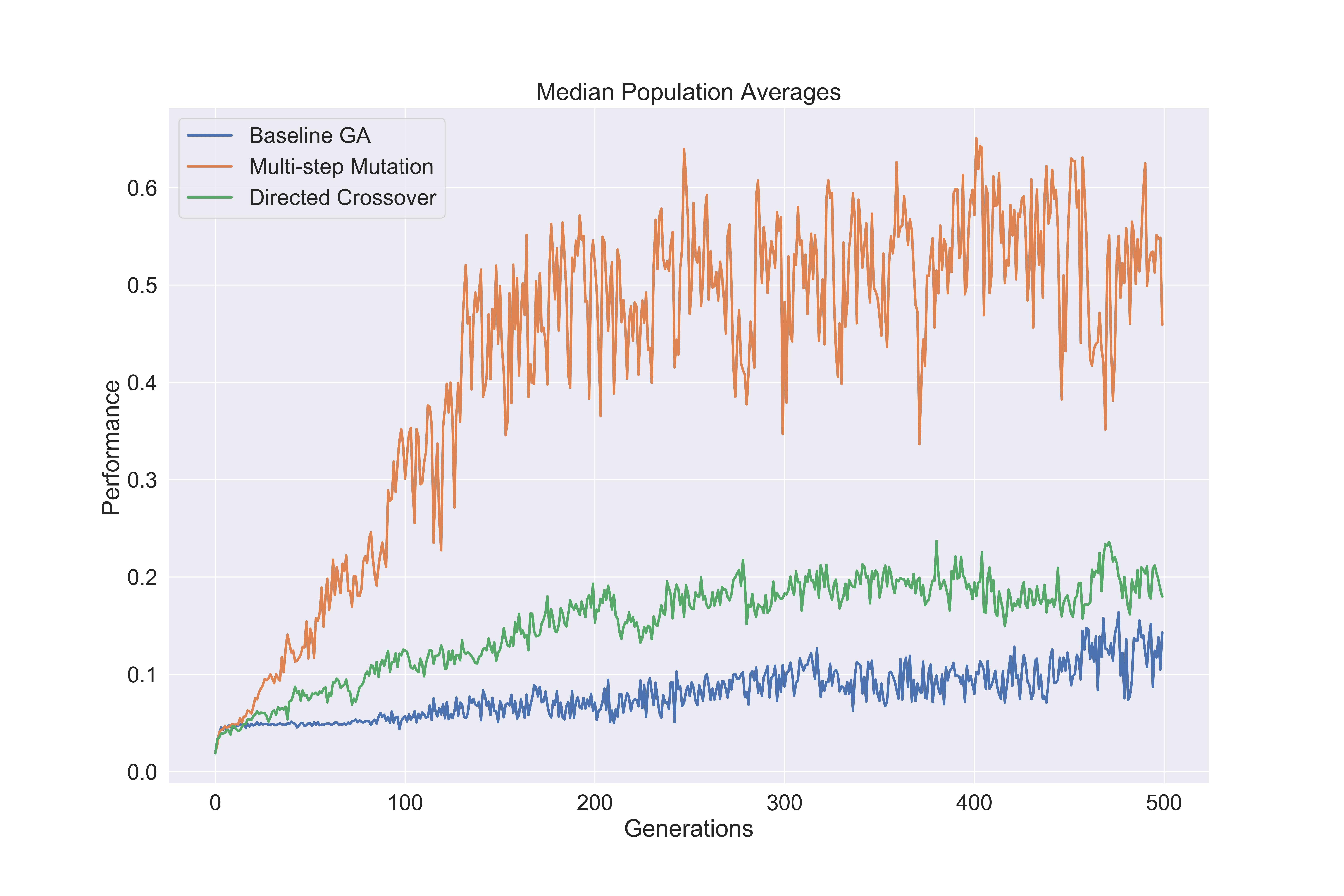}
\caption{Average Population}
\end{subfigure}
\begin{subfigure}{0.5\textwidth}
\includegraphics[width=\linewidth]{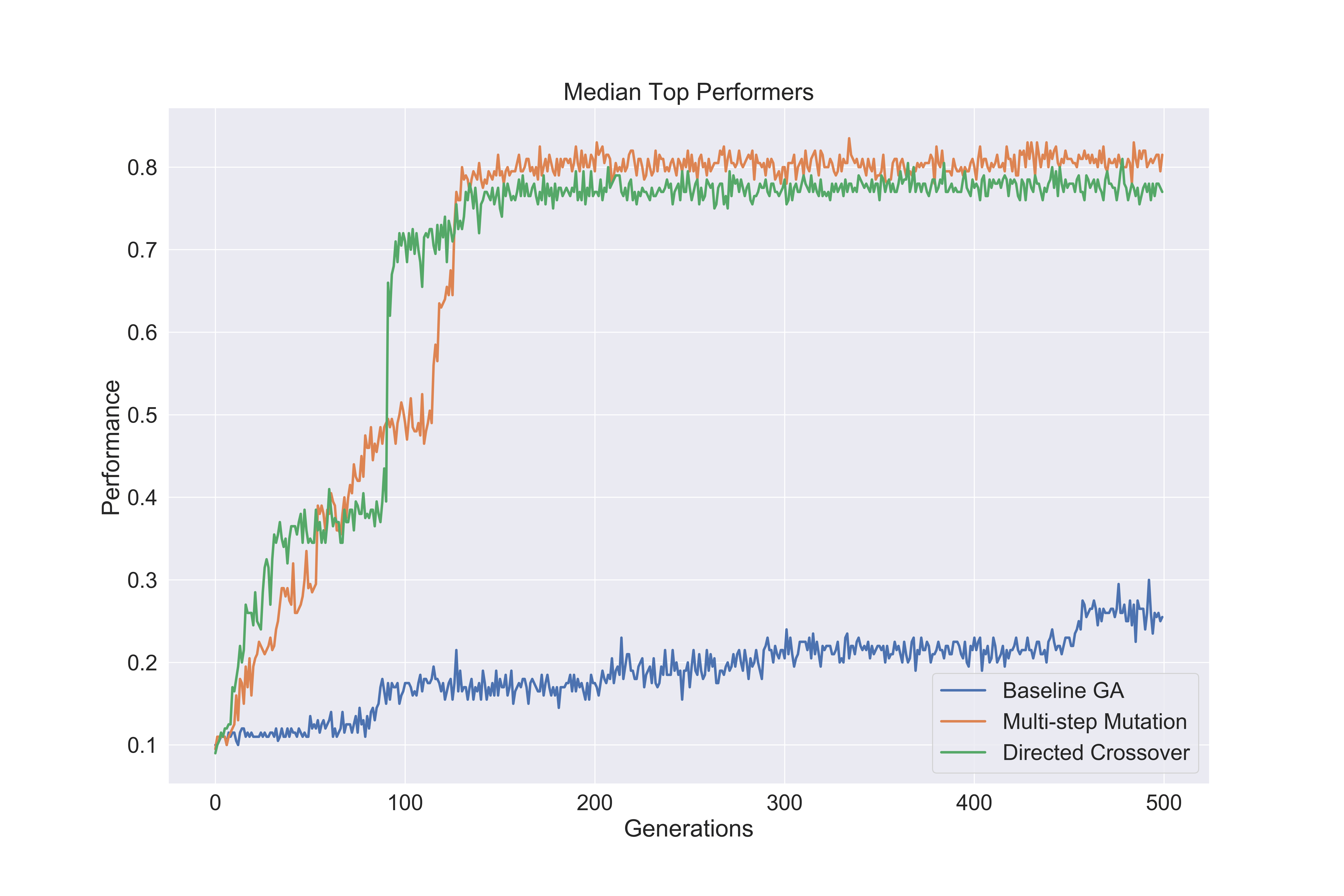}
\caption{Top Performers}
\label{fig:subim2}
\end{subfigure}
\caption[FrozenLake Neuroevolution Results]{FrozenLake Neuroevolution Results: The baseline implementation often takes longer than 500 generations to converge. MSM and Directed Crossover reliably speed up the process, typically producing optimal agents in under 150 generations.}
\label{fig:fl_results}
\end{figure}

For baseline neuroevolution and its modifications, two sets of measurements were made. We record the top performer's score in every generation in addition to the average score of the entire population, see \fig{fig:fl_results}. This score represents accumulated rewards over 100 episodes of the game. Ten samples of each variation were collected and the median scores are reported. The Mann-Whitney U score was computed between the baseline implementation and each of MSM and Directed Crossover for both the mean population score and the top performers. In the top performers' case, both MSM and Directed Crossover are significant ($p < 0.01$). There was no significant difference in population averages between Directed Crossover and the baseline, but MSM was significantly better ($p < 0.01$). This is not to say that the baseline fails in every case, the spread of baseline results is wide, and further testing showed that it does converge on an optimal solution, but it takes longer than the proposed modifications, which is the key difference.



\section{Discussion}
\label{S:6}

We presented two modifications to mutation and crossover in GAs and test their performance in a reinforcement learning environment. Both MSM and Directed Crossover reliably produce successful solutions in fewer generations than the baseline GA. One noteworthy difference is that the average population scores produced by MSM was higher than those produced by Directed Crossover or the baseline. This could be driven by the fact that the population is sometimes replaced altogether with one that uses euclidean distance from bad solutions as fitness and not FrozenLake performance. Going forward, it would be interesting to see how these methods do when combined since they alter different parts of the GA process.  

\bibliographystyle{plain}
\bibliography{references.bib}

\newpage
\appendix

\section{Spread of Results}
\label{Spread of Results}

\begin{figure}[h]
\centering
\includegraphics[width=0.48\textwidth]{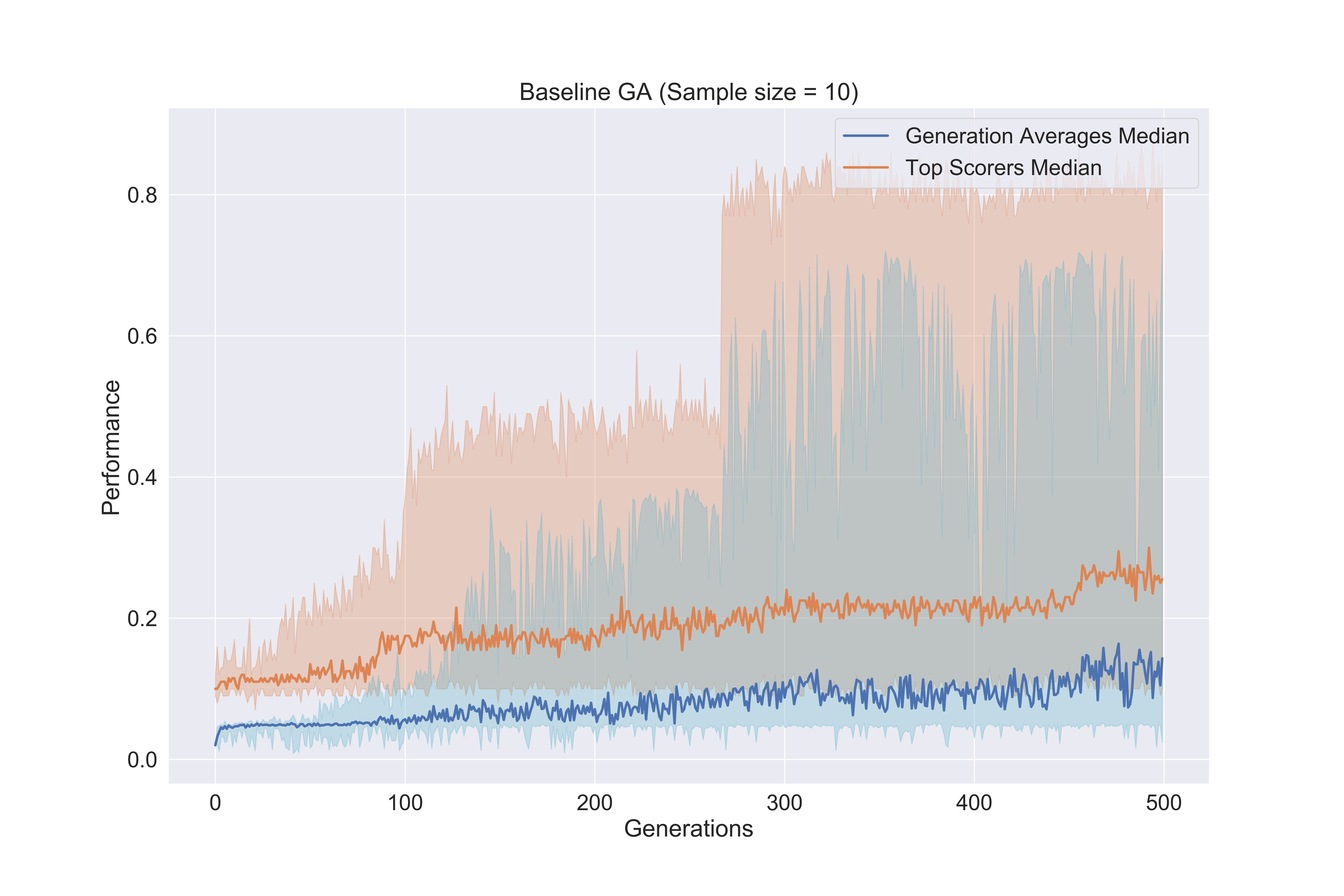}
\caption{Spread of baseline GA results}
\end{figure}

\begin{figure}[h]
\centering
\includegraphics[width=0.48\textwidth]{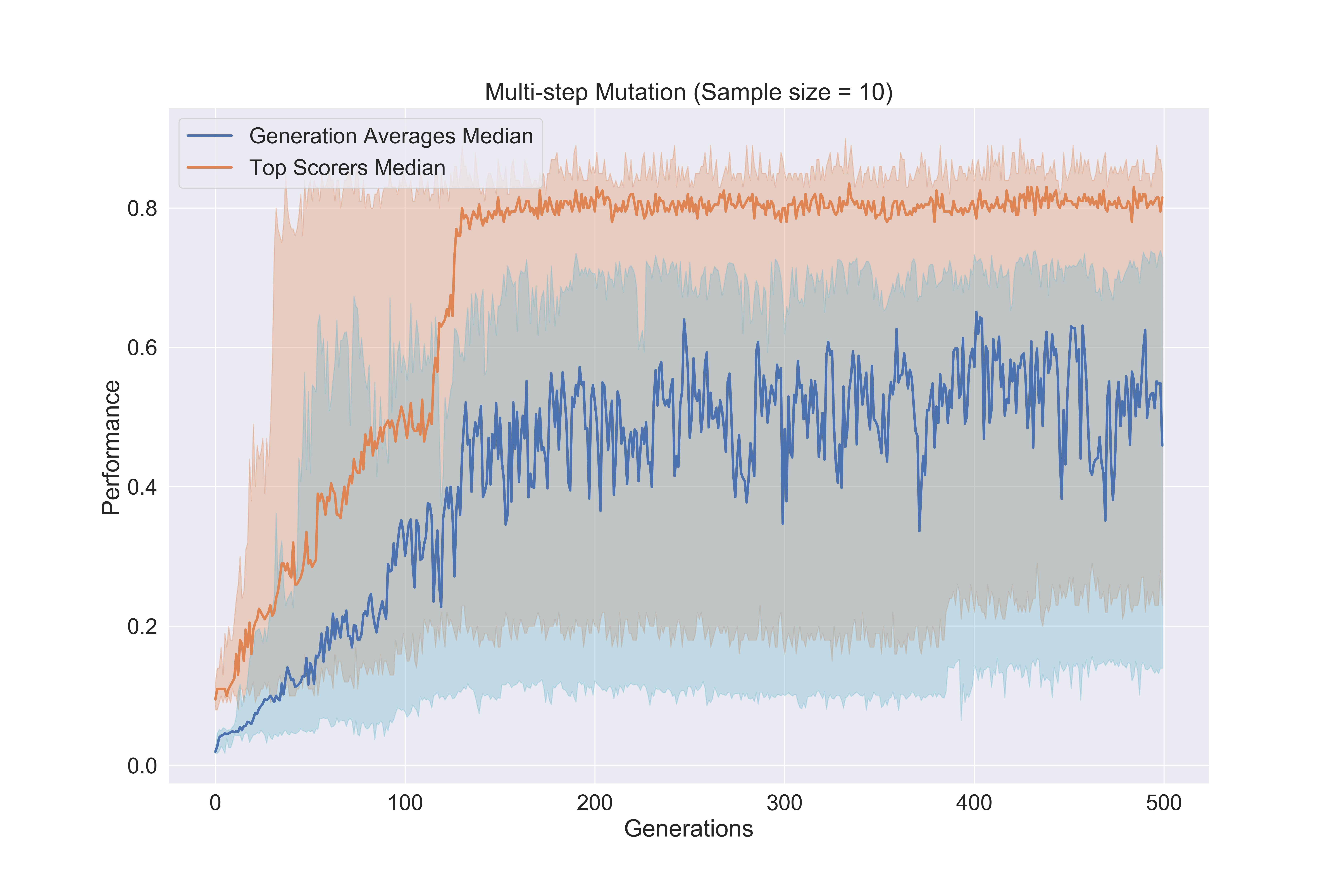}
\caption{Spread of multi-step mutation results}
\end{figure}

\begin{figure}[h]
\centering
\includegraphics[width=0.48\textwidth]{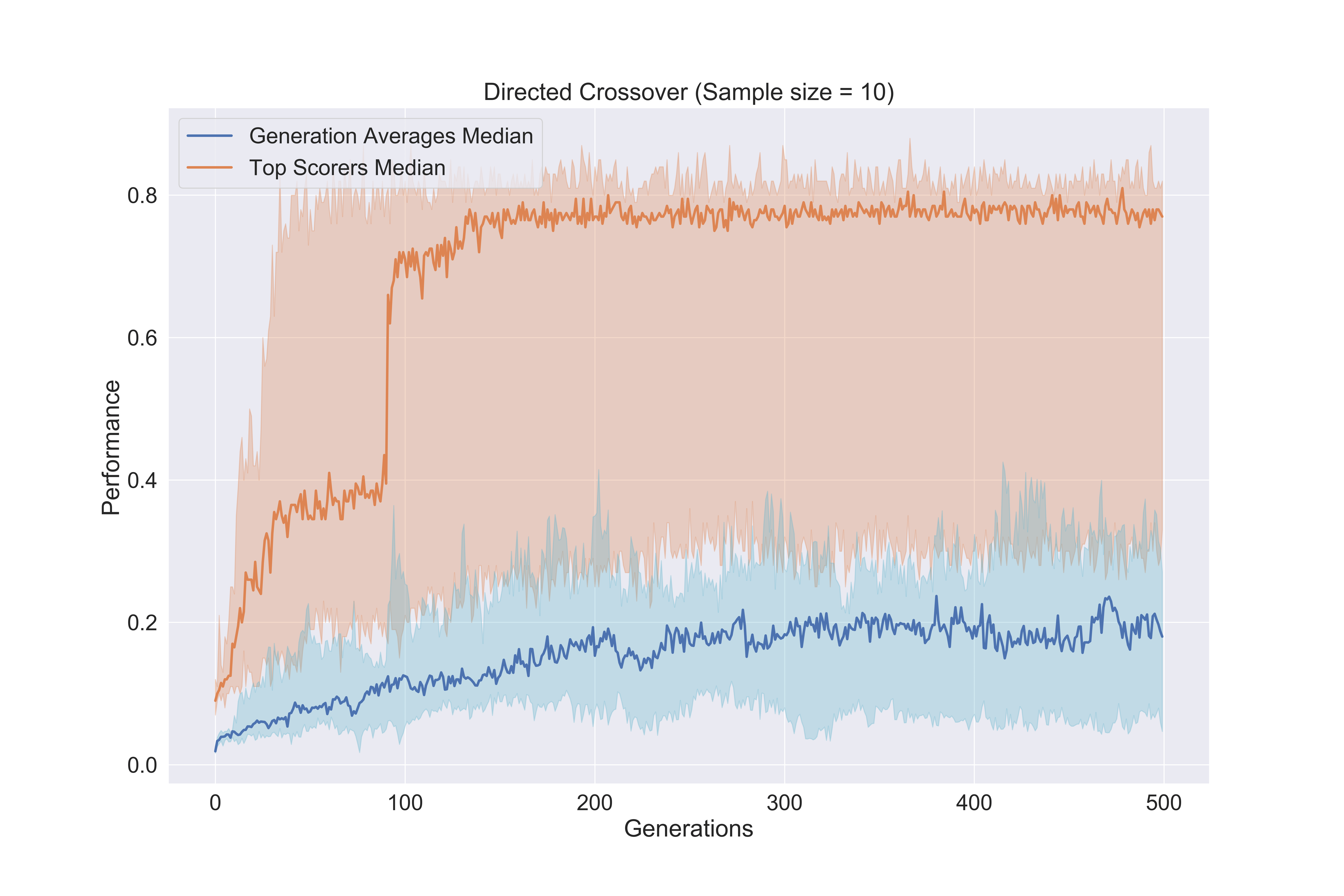}
\caption{Spread of directed crossover results}
\end{figure}

\section{Hyperparameters}
\label{Hyperparameters}
\begin{table*}[t]

\small
\centering
\begin{tabular}{ >{\centering\arraybackslash}m{3cm}  >{\arraybackslash}m{3cm}  >{\arraybackslash}m{6cm}}
\hline\hline

\textbf{GA} &  & Notes\\  
\hline
\hspace{1cm} & \hspace{1cm} & \hspace{1cm}\\
Network Type & Feed-Forward & -\\ 
\hspace{0.5cm} & \hspace{0.5cm} & \hspace{0.5cm}\\
\hline
\hspace{1cm} & \hspace{1cm} & \hspace{1cm}\\
Network Shape & 16-10-10-4 & Sixteen possible states as input, Four available actions as output, two hidden layers\\ 
\hspace{0.5cm} & \hspace{0.5cm} & \hspace{0.5cm}\\
\hline
\hspace{1cm} & \hspace{1cm} & \hspace{1cm}\\
Total Parameters & 300 & Number of weights in the network\\ 
\hspace{0.5cm} & \hspace{0.5cm} & \hspace{0.5cm}\\
\hline
\hspace{1cm} & \hspace{1cm} & \hspace{1cm}\\
Selection  & Elitist & Leaves the top 10\% of performers unchanged\\ 
\hspace{0.5cm} & \hspace{0.5cm} & \hspace{0.5cm}\\
\hline
\hspace{1cm} & \hspace{1cm} & \hspace{1cm}\\
Crossover  & Uniform or \newline Directed & Uses two parents, randomly chosen from \emph{elite} population\\ 
\hspace{0.5cm} & \hspace{0.5cm} & \hspace{0.5cm}\\
\hline
\hspace{1cm} & \hspace{1cm} & \hspace{1cm}\\
Mutation  & Gaussian \newline noise & Generated in the same shape as the parameter vectors and added to each individual with variance controlled by a decaying factor\\ 
\hspace{0.5cm} & \hspace{0.5cm} & \hspace{0.5cm}\\
\hline
\hspace{1cm} & \hspace{1cm} & \hspace{1cm}\\
MSM steps  & 10 & This is the number of extra times the elite population is crossed and mutated in the modified GA\\ 
\hspace{0.5cm} & \hspace{0.5cm} & \hspace{0.5cm}\\
\hline
\hspace{1cm} & \hspace{1cm} & \hspace{1cm}\\
$\zeta$  & 0.3 & Directed Crossover parameter that dictates the fraction of weights removed during crossover, used in a similar manner to \cite{Mocanu2018SET}\\ 
\hspace{0.5cm} & \hspace{0.5cm} & \hspace{0.5cm}\\
\hline

\end{tabular}
\label{table:nonlin}
\end{table*}









\end{document}